\documentclass[sigconf]{acmart}
\usepackage{balance}

\AtBeginDocument{%
  }

\copyrightyear{2024}
\acmYear{2024}
\setcopyright{rightsretained}
\acmConference[MM '24]{Proceedings of the 32nd ACM International Conference on Multimedia}{October 28-November 1, 2024}{Melbourne, VIC, Australia}
\acmBooktitle{Proceedings of the 32nd ACM International Conference on Multimedia (MM '24), October 28-November 1, 2024, Melbourne, VIC, Australia}
\acmDOI{10.1145/3664647.3681042}
\acmISBN{979-8-4007-0686-8/24/10}

\makeatletter
\gdef\@copyrightpermission{
  \begin{minipage}{0.3\columnwidth}
   \href{https://creativecommons.org/licenses/by/4.0/}{\includegraphics[width=0.90\textwidth]{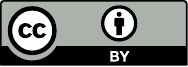}}
  \end{minipage}\hfill
  \begin{minipage}{0.7\columnwidth}
   \href{https://creativecommons.org/licenses/by/4.0/}{This work is licensed under a Creative Commons Attribution International 4.0 License.}
  \end{minipage}
  \vspace{5pt}
}
\makeatother

\begin{document}

\title[SynopGround: A Large-Scale Dataset for Multi-Paragraph Video Grounding from TV Dramas and Synopses]{SynopGround: A Large-Scale Dataset for Multi-Paragraph \\ Video Grounding from TV Dramas and Synopses}

\author{Chaolei Tan$^{*\dag}$}
\affiliation{%
  \institution{Sun Yat-sen University}
  \city{Guangzhou}
  \country{China}
}
\email{tanchlei@mail2.sysu.edu.cn}

\author{Zihang Lin$^{*\dag}$}
\affiliation{%
  \institution{Sun Yat-sen University}
  \city{Guangzhou}
  \country{China}
}
\email{linzh59@mail2.sysu.edu.cn}

\author{Junfu Pu}
\affiliation{%
  \institution{ARC Lab, Tencent PCG}
  \city{Shenzhen}
  \country{China}}
\email{jevinpu@tencent.com}

\author{Zhongang Qi}
\affiliation{%
  \institution{ARC Lab, Tencent PCG}
  \city{Shenzhen}
  \country{China}}
\email{zhongangqi@tencent.com}

\author{Wei-Yi Pei}
\affiliation{%
  \institution{Tencent Video, PCG}
  \city{Shenzhen}
  \country{China}}
\email{weiyipei@tencent.com}

\author{Zhi Qu}
\affiliation{%
  \institution{Tencent Video, PCG}
  \city{Shenzhen}
  \country{China}}
\email{jessonqu@tencent.com}

\author{Yexin Wang}
\affiliation{%
  \institution{Tencent Video, PCG}
  \city{Shenzhen}
  \country{China}}
\email{yexinwang@tencent.com}

\author{Ying Shan}
\affiliation{%
  \institution{ARC Lab, Tencent PCG}
  \city{Shenzhen}
  \country{China}}
\email{yingsshan@tencent.com}

\author{Wei-Shi Zheng}
\affiliation{%
  \institution{Sun Yat-sen University, Key Laboratory of Machine Intelligence and Advanced Computing, Ministry of Education}
  \city{Guangzhou}
  \country{China}}
\email{wszheng@ieee.org}

\author{Jian-Fang Hu$^{1\ddag}$}
\affiliation{%
  \institution{Sun Yat-sen University, Guangdong Province Key Laboratory of Information Security Technology}
  \city{Guangzhou}
  \country{China}}
\email{hujf5@mail.sysu.edu.cn}

\makeatletter
\def\authornotetext#1{
\if@ACM@anonymous\else
    \g@addto@macro\@authornotes{
    \stepcounter{footnote}\footnotetext{#1}}
\fi}
\makeatother
\authornotetext{Equal contribution.}
\authornotetext{Work done during an internship at ARC Lab, Tencent PCG.}
\authornotetext{Corresponding author.}

\def\authors{Chaolei Tan, Zihang Lin, Junfu Pu, Zhongang Qi, Wei-Yi Pei, Zhi Qu, Yexin Wang, Ying Shan, Wei-Shi Zheng, Jian-Fang Hu}

\renewcommand{\shortauthors}{Chaolei Tan et al.}

\begin{abstract}
Video grounding is a fundamental problem in multimodal content understanding, aiming to localize specific natural language queries in an untrimmed video. However, current video grounding datasets merely focus on simple events and are either limited to shorter videos or brief sentences, which hinders the model from evolving toward stronger multimodal understanding capabilities. To address these limitations, we present a large-scale video grounding dataset named SynopGround, in which more than 2800 hours of videos are sourced from popular TV dramas and are paired with accurately localized human-written synopses. Each paragraph in the synopsis serves as a language query and is manually annotated with precise temporal boundaries in the long video. These paragraph queries are tightly correlated to each other and contain a wealth of abstract expressions summarizing video storylines and specific descriptions portraying event details, which enables the model to learn multimodal perception on more intricate concepts over longer context dependencies. Based on the dataset, we further introduce a more complex setting of video grounding dubbed Multi-Paragraph Video Grounding (MPVG), which takes as input multiple paragraphs and a long video for grounding each paragraph query to its temporal interval. In addition, we propose a novel Local-Global Multimodal Reasoner (LGMR) to explicitly model the local-global structures of long-term multimodal inputs for MPVG. Our method provides an effective baseline solution to the multi-paragraph video grounding problem. Extensive experiments verify the proposed model's effectiveness as well as its superiority in long-term multi-paragraph video grounding over prior state-of-the-arts. Dataset and code are publicly available. Project page: \href{https://synopground.github.io/}{https://synopground.github.io/}.
\end{abstract}

\vspace{-15mm}



\maketitle
\vspace{1mm}
\section{Introduction}
\label{sec:intro}
\begin{figure}[t]
    \centering
    \includegraphics[width=1\linewidth]{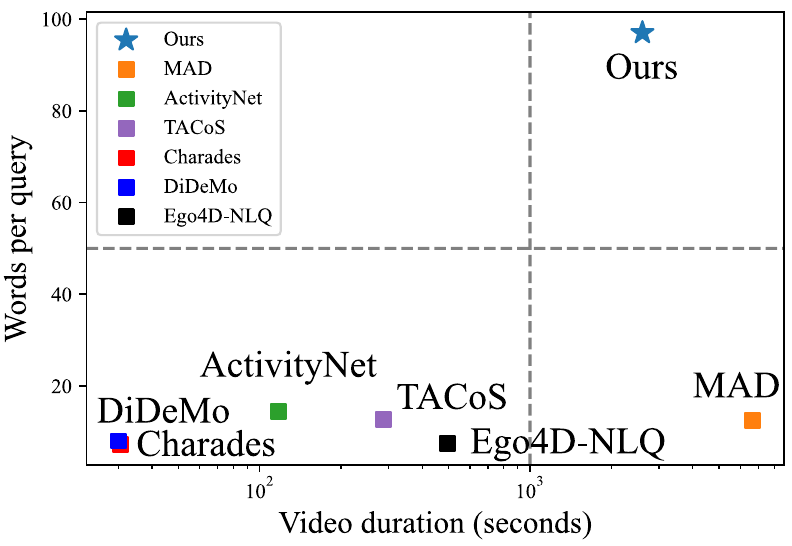}
    \vspace{-7.5mm}
    \caption{Comparison of video grounding datasets. Our SynopGround is the first dataset to introduce the challenges of both long videos and long queries into video grounding.}
    \label{fig:dataset_compare}
    \vspace{-6mm}
\end{figure}
As a crucial problem in multimodal understanding, video grounding aims at linking semantically relevant temporal intervals in an untrimmed video with specific natural language queries. Recently, video grounding has received increasing attention since a wide range of downstream applications can be promoted by it, such as improving the searching granularity of video retrieval~\cite{vr_1, vr_2, vr_3, vr_4, vr_5}, enabling language-aware scenarios of video editing~\cite{video_edit_1, video_edit_2, video_edit_3, video_edit_4}, and making video question answering~\cite{vqa_1, vqa_2, vqa_3, vqa_4, vqa_5, vqa_6, vqa_7} more evidence-based. Up to now, a large number of datasets and methods have been established to advance this line of research.

However, considerable drawbacks still exist in previous video grounding datasets~\cite{tall_charades, didemo, acitivitynet_caption, Tacos, ego4d, soldan2022mad}. First of all, as presented in Figure~\ref{fig:dataset_compare}, most of commonly-used datasets are constructed upon short videos and brief sentence queries. This setup limits the model in developing stronger abilities that can model and bridge the long-form videos~\cite{cone, scan_once} and long-text queries~\cite{subquery, tall_charades}. Besides, shorter queries that describe detailed events (as shown in Table \ref{tab:query_compare}), are more prone to causing the risk of semantic ambiguity in referring expressions~\cite{refcocog, uncovering}, i.e., the occurrence of one-to-many correspondence between queries and moments, which will adversely affect the model learning. In particular, this ambiguity issue is more prominent for the recently proposed MAD~\cite{soldan2022mad} dataset which features long input videos but short general descriptions. For example, it is highly likely to find cases where multiple moments are semantically corresponding to the same short description like ``She steps closer.'' (shown in Table~\ref{tab:query_compare}), especially when searching content in a long video. Furthermore, as listed in Table~\ref{tab:query_compare}, existing benchmarks are tailored for language queries referring to low-level visible activities, while all of them overlook the importance of more complex events and abstract concepts. Such drawbacks actually limit the applications of video grounding in scenarios where complex descriptions with abstract concepts should be associated with long-term videos. For example, accelerating the movie post-production by automatically integrating raw footage into a coherent story based on the plot scripts is a practical need, but it cannot be satisfied by the current video grounding techniques developed from existing datasets.

In this work, we curate and present a large-scale dataset called SynopGround to address the current limitations of video grounding datasets. We collect and manually annotate episodes from popular TV dramas of various genres, yielding a large-scale video grounding dataset consisting of over 2800 hours of fully-annotated videos. Specifically, for each video, we crawl its human-written synopsis consisting of multiple paragraphs from the Internet, and further annotate the precise temporal boundaries for each paragraph in the given synopsis. As demonstrated in Figure~\ref{fig:dataset_compare}, our dataset has both significantly longer average video length and average query length than most existing ones. It is the first video grounding dataset that can support the research on long-term contextual video grounding with complex queries. Moreover, compared to the short sentence queries in the existing datasets, our long paragraph queries can unambiguously indicate one-to-one correspondence between language queries and target moments, which is crucial for learning accurate cross-modal alignment. Furthermore, as shown in Table~\ref{tab:query_compare}, there are very concrete descriptions for visible activity concepts like ``go to the cabin'', as well as extremely concise and abstract expressions like ``spent a happy time'' in the query from our dataset. This enables to learn and evaluate the comprehensive understanding of semantic concepts at diverse abstraction levels.
\begin{table}[t]
    \centering
    \caption{Comparison of queries in different datasets. The red-bold text is a detailed description, while the blue-italic text is an abstract and concise expression.}
    \vspace{-3mm}
    \footnotesize
    \begin{tabular}{|c|p{5.5cm}|}
    \hline
    \textbf{Dataset} & \textbf{Query} \\ \hline
    Charades~\cite{tall_charades} & A person runs to the window then looks out. \\ \hline
    DiDeMo~\cite{didemo} & The little girl jumps back up after falling. \\ \hline
    TACoS~\cite{Tacos} & He flips the eggs, making an omelet. \\ \hline
    ActivityNet\cite{acitivitynet_caption} & A woman walks to the piano and briefly talks to the elder man. \\ \hline
    Ego4d-NLQ~\cite{ego4d} & What did I pick from the fridge? \\ \hline
    MAD~\cite{soldan2022mad} & She steps closer. \\ \hline
    SynopGround (Ours) & \ldots Stefan and Elena decided to {\color{red}\textbf{go to the cabin}} left by Elena's parents, where they {\color{blue}\textit{spent a happy time.}} \space \ldots \\ \hline
    \end{tabular}
    \label{tab:query_compare}
    \vspace{-6mm}
\end{table}

Based on our dataset, we pioneer to introduce and explore a more challenging and complex setting of video grounding called Multi-Paragraph Video Grounding (MPVG). The MPVG task receives a multi-paragraph synopsis and a long narrative video as inputs to localize the temporal interval of each synopsis paragraph from the video. To promote and inspire future research, we further propose a novel Local-Global Multimodal Reasoner (LGMR) to explicitly model the local-global structures of long-term multimodal inputs and conduct iterative cross-modal reasoning within and across the two levels of structures for effectively tackling the multi-paragraph video grounding problem. Extensive experiments demonstrate the effectiveness of our baseline in the proposed research direction.

The main contributions of this work are summarized as follows:
\begin{itemize}
    \item We present SynopGround, a large-scale video grounding dataset consisting of over 2800 hours of TV drama videos with manual temporally-annotated professional synopses.
    \item Based on the dataset, we first introduce a challenging Multi-Paragraph Video Grounding (MPVG) task and propose a novel Local-Global Multimodal Reasoner (LGMR) baseline.
    \item We are the first to incorporate long-form videos with long abstract paragraphs for video grounding. Comparison results show the unique advantages of our dataset and the efficacy of our baseline in multi-paragraph video grounding.
\end{itemize}

\section{Related Work}
\begin{figure*}[t]
    \centering
    \includegraphics[width=1.0\linewidth]{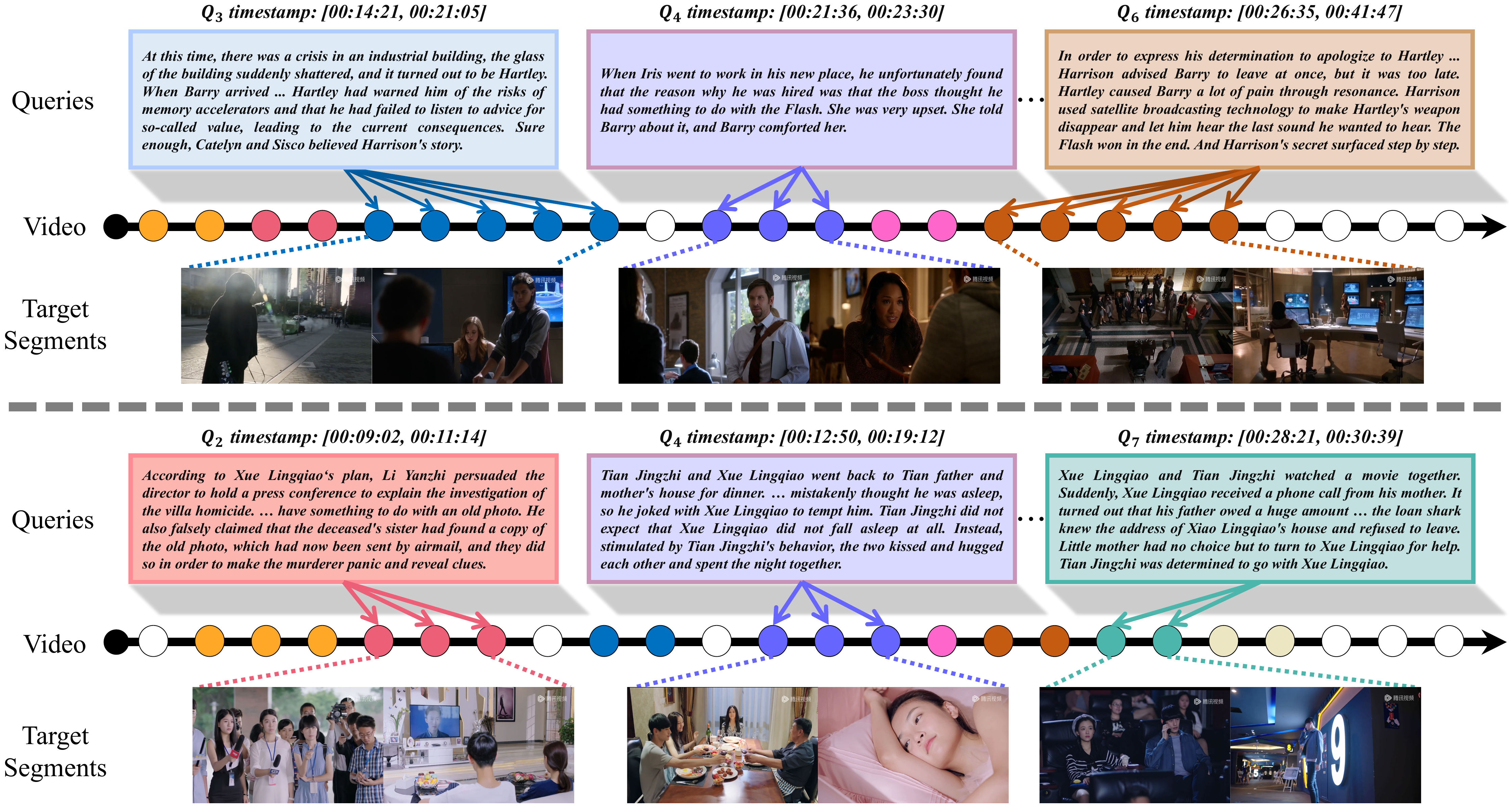}
    \vspace{-6mm}
    \caption{Illustration of the proposed Multi-Paragraph Video Grounding (MPVG) problem and two representative samples in our SynopGround. Given a video and a synopsis $\mathcal{Q}$ that contains $N$ paragraphs $\{Q_1, Q_2, ..., Q_N\}$, the model should predict the corresponding temporal interval for each paragraph $Q_i$ in the form of starting and ending time.}
    \label{fig:sample}
    \vspace{-3mm}
\end{figure*}
In this section, we aim to review and discuss the existing works in the video grounding and narrative video understanding areas.
\subsection{Video Grounding}
\vspace{1mm}
\noindent\textbf{Datasets.}
In video grounding, Charades-STA~\cite{tall_charades}, DiDeMo~\cite{didemo}, ActivityNet-Captions~\cite{acitivitynet_caption} and TACoS~\cite{Tacos} are the four most commonly used datasets for model training and evaluation. However, a majority of these datasets~\cite{tall_charades, acitivitynet_caption, Tacos} are adapted from pre-existing video datasets tailored for closed-set recognition or localization tasks~\cite{charades_ori, anet_ori, tacos_ori}, which makes them severely limited to a pre-defined set of visual and linguistic concepts. DiDeMo~\cite{didemo} is a customized video grounding dataset. However, it overly simplifies the annotation and only supports the model to select from 5 evenly-divided segments of the video. In addition, shortcut learning issues caused by distribution biases in previous datasets have been reported~\cite{uncovering, closerlook}, which could adversely affect the benchmark reliability. Moreover, the above datasets~\cite{tall_charades, didemo, acitivitynet_caption, Tacos} are all constructed on a relatively small-scale collection of short videos and simple sentence descriptions, which cannot support the need of large-scale model training for long-term contextual video-language understanding that incorporates complex language queries. Recently, the Ego4d-NLQ~\cite{ego4d} and MAD~\cite{soldan2022mad} datasets are introduced. Nevertheless, both of them still focus on the simple visible activities and short-term temporal events. Specifically, Ego4d-NLQ contains egocentric videos and adopts brief interrogative queries asking about simple visible fact grounded on a short video interval. The MAD dataset is semi-automatically constructed on movies with audio descriptions and its average video length is significantly longer compared to the other existing datasets. However, the language queries of MAD are still brief sentences that individually describe short-term events in the long video. Different from all of the prior works, our proposed SynopGround is the first video grounding dataset that considers both long-form videos and long-text queries. Additionally, we adopt narrative videos conveying storylines and tightly correlated synopsis paragraphs as inputs, which poses more challenges for the video grounding model to understand high-level story plots and invisible abstract concepts over a longer context.
\begin{figure*}[t]
    \centering
    \includegraphics[width=\linewidth]{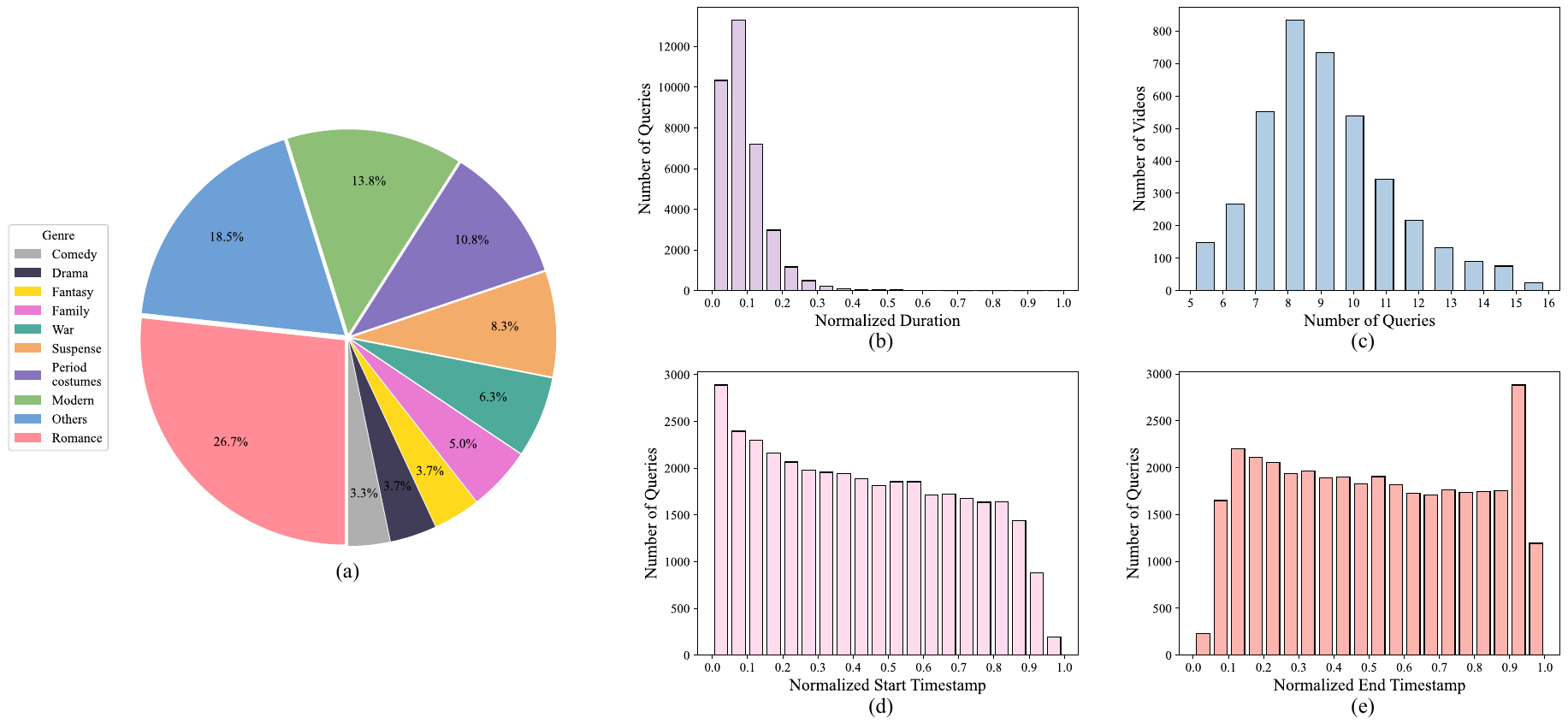}
    \vspace{-6mm}
    \caption{Data distribution in our dataset. (a): Genre distribution of TV dramas. (b): Normalized duration of target video segments. (c): Number of queries per video. (d): Normalized start timestamp distribution. (e): Normalized end timestamp distribution.}
    \label{fig:distribution}
    \vspace{-0mm}
\end{figure*}

\noindent\textbf{Tasks and settings.} Early research of video grounding has largely focused on grounding single sentences in videos, i.e., the Video Sentence Grounding (VSG) task introduced by Gao et al.~\cite{tall_charades} and Hendricks et al.~\cite{didemo}. Afterwards, a series of extended tasks~\cite{vcmr_task, tvr, qvhighlight, depnet} have been proposed. Escorcia et al.~\cite{vcmr_task} first introduced the task of Video Moment Corpus Retrieval (VCMR) for combining video retrieval and moment localization, and Lei et al.~\cite{tvr} curated the TVR dataset to incorporate multi-modal information into VCMR. Lei et al. ~\cite{qvhighlight} proposed QVHighlights dataset along with a new direction combining highlight detection and moment retrieval. To reduce ambiguity by exploring inter-query context, some recent works~\cite{depnet, svptr, prvg, wsag, hscnet, htstep} have shifted to a multi-query version of video sentence grounding, where the model is required to understand several temporally ordered sentences and localize each sentence in a richer context. Specifically, Bao et al. ~\cite{depnet} first studied multi-sentence video grounding in a fully-supervised setting, and the semi-supervised setting~\cite{svptr} as well as weakly-supervised setting~\cite{siamgtr} have also been investigated after that. These prior works have shown the great potential of contextually understanding multimodal content in untrimmed videos and language descriptions. In this work, we take a step further to introduce a more challenging setting of contextual video grounding called Multi-Paragraph Video Grounding (MPVG). It requires to understand both short-term intra-paragraph semantics and long-term inter-paragraph dependencies, which connects the complex temporal structures of long videos with the complicated semantics of long paragraphs.

\noindent\textbf{Methods.} A lot of approaches~\cite{ablr, acmmm_2019, 2dtan, drn, lgi, acmmm_2020, vslnet, man, ivg, debug, acmmm_2021, rethinkingbu, bpnet, cbln, mmvtg, mmn, valuncertainty, acmmm_2022, can, where2focus, g2l, acmmm_2023, aug-2dtan, csdvl} have been developed for video grounding over recent years. As summarized in ~\cite{tsgv_survey}, these methods can be roughly categorized into proposal-based and proposal-free methods. Proposal-based methods typically involve a two-stage process of generating moment proposals for relevance score ranking, which often leads to issues like inefficiency and limited adaptability. In contrast, proposal-free methods tend to have better efficiency and they directly predict the temporal boundaries based on cross-modal interactions, which is more suitable for various real-world scenarios. Considering the long-term characteristics in our dataset, we choose to design our Local-Global Multimodal Reasoner (LGMR) in the more efficient proposal-free fashion. Different from the previous approaches that only consider modeling the cross-modal correspondence between a single paragraph and the video, our method is constructed by reasoning through the local and global structures of multiple paragraphs and the video.

\vspace{-2mm}
\subsection{Narrative Video Understanding}
Understanding visual content presented in narrative videos is an important area with many works~\cite{simoes2016movie, haurilet2016naming, nagrani2018benedict, huang2018unifying, gu2018ava, del2013state, huang2020movienet, rao2020unified, tapaswi2016movieqa, jasani2019we, soldan2022mad, xiong2019graph, bain2020condensed, sun2022synopses} proposed accordingly. Most of these prior works neglect to model and understand the content of narrative videos based on their high-level storylines while focusing on specific downstream applications, such as movie genre classification~\cite{simoes2016movie}, character identification~\cite{haurilet2016naming, nagrani2018benedict, huang2018unifying}, action localization~\cite{gu2018ava}, scene segmentation~\cite{del2013state, huang2020movienet}, and shot classification~\cite{rao2020unified}. In addition to that, some works have started pursuing story-level understanding in many different ways. For instance, Tapaswi et al. ~\cite{tapaswi2016movieqa} proposed MovieQA dataset to comprehend movie stories by question-answering. MSA~\cite{xiong2019graph}, CMD~\cite{bain2020condensed}, and SyMoN~\cite{sun2022synopses} datasets utilized synopses as language queries and formulated movie understanding as text-to-video retrieval. This line of research is closely related to ours. However, we focus on a more challenging video grounding task with long contexts and complex queries, in which the model should understand the long-range cross-modal dependencies so as to reason about the video grounding results at story level.

\vspace{-1.5mm}
\section{SynopGround Dataset}
\vspace{0.5mm}
\begin{table*}[h]
    \centering
    \renewcommand{\arraystretch}{1.25}
    \caption{Detailed comparison with existing video grounding datasets. Our SynopGround is at a larger scale in terms of the total duration of videos and it contains precise temporal annotations generated by human annotators. It is also the first large-scale dataset that considers both long-form videos and long-text queries for multi-paragraph video grounding.}
    \vspace{-3mm}
   \resizebox{!}{23.5mm}{
   \begin{tabular}{c|c|c|c|c|c|c|c}
      \hline
      Dataset & Charades-STA\cite{tall_charades} & ANet-Captions\cite{acitivitynet_caption} & DiDeMo\cite{didemo} & TACoS\cite{Tacos} & Ego4d-NLQ\cite{ego4d} & MAD\cite{soldan2022mad} & \textbf{SynopGround} \\ 
      \hline
      Domain & Indoor & Open & Open & Cooking & Open & Open & Open \\
      \hline
      Annotation Mode & Semi-Automatic & Manual & Manual & Manual & Manual & Semi-Automatic & Manual \\
      \hline
      Paragraph Query & No & No & No & No & No & No & \textbf{Yes} \\
      \hline
      \# Videos & 6,672 & \textbf{14,926} & 10,464 & 127 & 1659 & 650 & 3,987\\
      \hline
      \# Queries & 16,124 & 71,953 & 40,543 & 18,818 & 19,170 & \textbf{384,600} & 36,002 \\
      \hline
      \# Words / Query & 7.2 & 14.4 & 8.0 & 12.7 & 7.5 & 12.4 & \textbf{97.0} \\
      \hline
      Duration / Video & 30.6s & 117.6s & 30.0s & 286.6s & 495s & \textbf{6,646.0s} & 2,608.4s \\
      \hline
      Duration / Moment & 8.1s & 37.1s & 6.5s & 6.1s & 3.9s & 4.1s & \textbf{239.5s} \\
      \hline
      Total Duration & 57.1h & 487.6h & 88.7h & 10.1h & 228.1h & 1,200.0h & \textbf{2,884.9h} \\
      \hline
   \end{tabular}}
    \vspace{-2mm}
   \label{tab:Comparison}
\end{table*}
\begin{table}[t]
    \centering
    \caption{Statistics of dataset division.}
    \vspace{-3mm}
    \resizebox{!}{9mm}{
    \begin{tabular}{l|c|c|c}
    \hline
    Data Split & \# Dramas & \# Videos & \# Queries \\
    \hline
    Training & 470 & 3,187 & 28,677 \\
    Validation & 190 & 400 & 3,791 \\
    Testing & 192 & 400 & 3,534 \\
    \hline
    \end{tabular}
    }
    \label{tab:splits}
    \vspace{-5mm}
\end{table}
In this section, our goal is to give a formal definition of our introduced multi-paragraph video grounding problem and illustrate the details of the data collection, annotation, statistics, and processing.

\subsection{Problem Formulation}
Considering video paragraph grounding~\cite{depnet} is limited to a multi-query version of short single sentence grounding, we introduce a more challenging setting to incorporate long abstract paragraphs as queries called Multi-Paragraph Video Grounding (MPVG). Specifically, given an untrimmed video $\mathcal{V}$ and $N$ consecutive paragraph queries $\mathcal{Q}=\{Q_1, Q_2, ..., Q_N\}$ as input, the output should be $N$ temporal intervals $\{\mathcal{T}_1, \mathcal{T}_2, ..., \mathcal{T}_N\}$ corresponding to each of the paragraph queries, where $\mathcal{T}_i = \left(t_s^{(i)}, t_e^{(i)}\right)$ indicates the starting timestamp $t_s^{(i)}$ and ending timestamp $t_e^{(i)}$ for the $i$-th paragraph query in the target video. In our dataset, the video $\mathcal{V}$ is an episode from a TV drama, and $\mathcal{Q}$ is the corresponding human-written synopsis that contains $N$ paragraphs, with $Q_i$ indicating the $i$-th paragraph in the synopsis $\mathcal{Q}$. Illustration of our MPVG is in Figure~\ref{fig:sample}.

\subsection{Data Collection and Annotation}
We collect all the TV drama episodes from a leading online platform Tencent Video with official acknowledgement and permission. The plot synopsis for each episode of the TV dramas is scraped from a specialized TV review website\footnote{URL: https://www.tvmao.com. All texts are translated into English using Tencent Cloud Translation for research purpose only.} that contains lots of synopses of the most popular TV drama episodes written by professionals. Synopses that are too long or too short are discarded to ensure an adequate number of paragraphs in each synopsis. A total of 520 licensed and high-viewership TV dramas with textual synopses are finally selected to constitute our dataset, and we randomly sample several episodes from each selected TV drama to further annotate. Specifically, annotators are asked to read and understand the synopsis in advance. They then thoroughly watch the corresponding TV drama episode to determine the starting and ending time of the video content depicted by each synopsis paragraph.

Our data annotation pipeline is organized into multiple rounds to ensure the annotation quality. Specifically, all collected videos are first divided into numerous disjoint subsets of videos. In each annotation round, synopsis paragraphs for videos in one subset will be annotated with timestamps and each annotator is told to provide a score to indicate level of confidence in the annotated results. Afterwards, we first discard samples with low confidence as an initial cleanup, and then some of the remaining samples are selected to be manually checked in terms of quality. If the annotation quality is thought of as satisfactory, the annotation process will move on to another unlabeled subset of video data. Otherwise, the current batch of data would be re-annotated. The above procedures are repeated by tens of annotators until we finish the annotation of all candidate samples. For post-annotation assessment, we randomly select a proportion of the annotated data to be re-annotated by other annotators. Concretely, we calculate the temporal IoU (Intersection over Union) between the two results from different annotators, which reaches a value of about 85\%. This assessment result is much better than those of other datasets, such as the ActivityNet~\cite{acitivitynet_caption}, where different annotators only achieve an agreement degree around 70\%. The higher degree of agreement across different annotators in our dataset verifies the effectiveness of our designed pipeline for data annotation and quality control.

\begin{figure*}[t]
    \centering
    \includegraphics[width=1.0\linewidth]{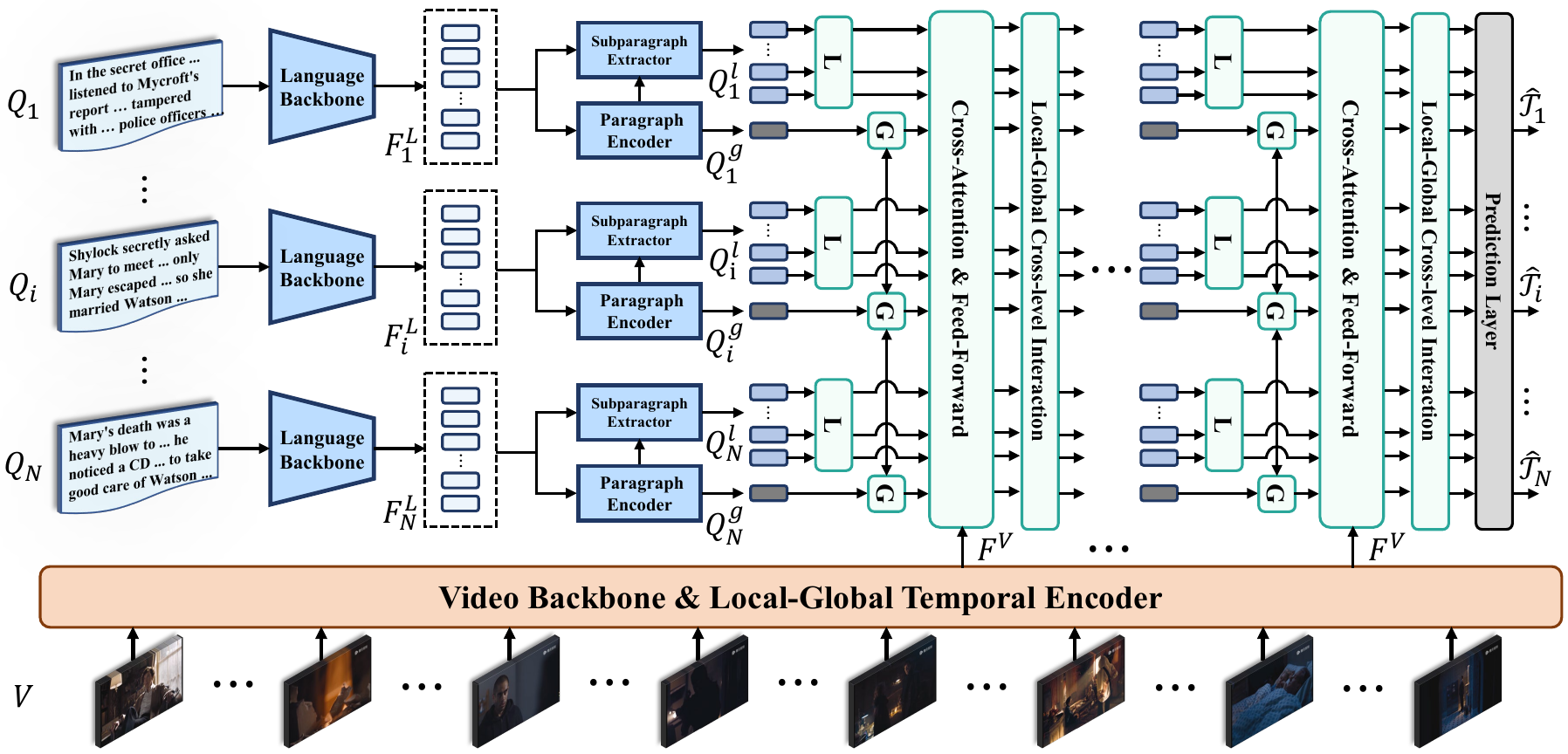}
    \vspace{-7.5mm}
    \caption{Our proposed Local-Global Multimodal Reasoner (LGMR). It consists of a local-global temporal encoder for structural long-term temporal modeling and a local-global iterative decoder to adaptively reason through local and global semantics.}
    \label{fig:framework}
    \vspace{-0mm}
\end{figure*}

\vspace{-2mm}
\subsection{Data Statistics}
\noindent\textbf{Data distribution.} We first illustrate some statistical distributions of our dataset in Figure~\ref{fig:distribution}. As shown in Figure~\ref{fig:distribution} (a), the TV dramas used in our dataset cover a wide spectrum of genres, which demonstrates the diversity of the collected data. In Figure~\ref{fig:distribution} (b), we show the normalized duration of the target video segments. Most of the target video segments cover less than 20\% of the full episode, which can be challenging for the model to correctly localize. In Figure~\ref{fig:distribution} (c), we visualize the distribution of the number of queries/paragraphs in each synopsis, and most synopses are composed of 5-13 paragraphs. Exploring the contextual information among these paragraphs is important for achieving promising performance in our multi-paragraph video grounding task. In Figure~\ref{fig:distribution} (d) and (e), we visualize the temporal distributions of the starting timestamps and ending timestamps of the target video segments. Both of them approximately present a uniform distribution, which ensures the model cannot benefit much from the distribution bias.

\noindent\textbf{Detailed comparison with other datasets.} In Table~\ref{tab:Comparison}, we compare our dataset with other existing datasets in detail. As suggested, our videos are much longer in duration than those of Charades-STA~\cite{tall_charades}, ActivityNet-Captions~\cite{acitivitynet_caption}, DiDeMo~\cite{didemo}, TACoS~\cite{Tacos} and Ego4d-NLQ~\cite{ego4d}. Although the average video duration in our dataset is shorter than that of MAD, our total duration of videos is more than twice that of MAD, showing that our dataset is at a larger scale. Furthermore, the duration of target segments in our dataset is significantly longer while the normalized target span is still short, making our target moments challenging to be localized. Note that some datasets like MAD have shorter normalized target span than ours, but their short and general queries bring the harmful and undesirable semantic ambiguity issue as mentioned before. In addition, our dataset is the first to incorporate paragraph queries, and the average number of words in each query is significantly larger than those of other datasets, which greatly reduces the semantic ambiguity of the queries. Moreover, our synopsis queries involve both abstract expressions and concrete descriptions, enabling the model to learn semantic concepts at more diverse abstraction levels.

\noindent\textbf{Data splits.}
As shown by the statistics in Table~\ref{tab:splits}, we carefully divide the entire data into three non-overlapping splits for training, validation, and testing. Our training, validation, and testing sets consist of 3,187, 400, and 400 videos, respectively. It is worth noting that each video is an episode from a TV drama. Additionally, we manually guarantee half of the videos for validation/testing are sourced from TV dramas that are not selected for the training set.
\vspace{-3mm}
\subsection{Data Pre-processing}
To promote data utilization, we provide pre-extracted features for public release. Specifically, pre-trained CLIP~\cite{clip} ViT-L/14 model is adopted to extract frame features for videos using a sampling rate of 3 FPS. Additionally, we extract segment features for videos with SlowFast~\cite{feichtenhofer2019slowfast} network, which is pre-trained on Kinectics-600~\cite{carreira2017quo, carreira2018short} and AVA~\cite{gu2018ava} datasets. To capture the character and dialogue information related to the storylines, we extract embedded subtitles using OCR models DBNet~\cite{liao2020real} and SVTR~\cite{du2022svtr}. Then, we encode each extracted subtitle to a feature representation using a pre-trained RoBERTa~\cite{liu2019roberta} model.
The pre-extracted CLIP, SlowFast, and OCR features describe the video from different aspects. They provide complementary information of the video and are beneficial for aligning synopsis with the video. Due to copyright restrictions, we cannot release the raw video frames but we will provide URL links where researchers can access and view the original videos.

\vspace{-1mm}
\section{Method}
In this section, we illustrate the details of our proposed baseline method to tackle the multi-paragraph video grounding problem.
\vspace{-3mm}
\subsection{Overview}
As shown in Figure~\ref{fig:framework}, our proposed Local-Global Multimodal Reasoner (LGMR) consists of a local-global temporal encoder for encoding the long input videos and a local-global iterative decoder for decoding the long paragraph queries. The video encoder decomposes the temporal correlations of long videos into intra-window and inter-window parts for efficient long-term temporal modeling. The query decoder first extracts subparagraph representations with a set of learnable queries guided by the global semantics of paragraphs, and then repeatedly conducts cross-modal reasoning within and across the local and global queries. We elaborate on more architectural details in the following.

\vspace{-1.5mm}
\subsection{Local-Global Temporal Encoder}
Given the long-form video inputs, we design a local-global attentive encoder to capture the evolving temporal dynamics of long-term video content, which exploits more structural temporal information than straightforward full attention. Specifically, we first project the video features of length $T$ into a hidden dimension of $D$, then for each video encoder layer, we split the input video feature sequence into non-overlapping temporal windows of length $M$, resulting in the intra-window video representations $F^{w}\in \mathbb{R}^{K\times M\times D}$, where $K$ is the total number of the temporal windows. For the video features in the $i$-th window, i.e., $F^{w}_{i}\in \mathbb{R}^{M\times D}$, we first encode the detailed local information by performing temporal self-attention within the scope of that window as follows:
\vspace{-1mm}
\begin{equation}
    F^{\ell}_{i} = \text{Self-Attention}\left(F^{w}_{i}, F^{w}_{i}, F^{w}_{i}\right)
    \vspace{-1mm}
\end{equation}
where $F^{\ell}_{i}\in \mathbb{R}^{M\times D}$ is the encoded video features with rich local contexts. Based on the local video features, we further apply a global-level self-attention on the global window features to connect different local contexts. Instead of using a simple pooling layer to aggregate the local window features, we exploit an attention-based method similar to the attention pooling in CLIP~\cite{clip} to dynamically gather important local contexts for global interaction as:
\vspace{-1mm}
\begin{equation}
    F^{g}_{i} = \text{Cross-Attention} \left( \text{Avg}\left(F^{l}_{i}\right), F^{l}_{i}, F^{l}_{i} \right)
    \vspace{-1mm}
\end{equation}
where $F^{g}\in \mathbb{R}^{K\times D}$ represents the global window features across the entire video, and $\text{Avg}\left(\cdot\right)$ is an average operation across temporal dimension for the local window features. Then the global window features are interacted with each other by a self-attention as:
\vspace{-1mm}
\begin{equation}
    F^{g} \leftarrow \text{Self-Attention}\left(F^{g}, F^{g}, F^{g}\right)
    \vspace{-1mm}
\end{equation}
Next, we merge the information from the local and global contexts of the intra-window and inter-window features as:
\vspace{-1mm}
\begin{equation}
    F^{V} = \text{FFN}\left(\text{LN}\left(\text{Flatten}\left(F^{\ell} + \text{Rep}\left(F^{g}\right)\right)\right)\right)
    \vspace{-1mm}
\end{equation}
where $\text{Rep}\left(\cdot\right)$ and $\text{Flatten}\left(\cdot\right)$ respectively indicate repeating the global window features by $M$ times in its corresponding window and unfolding the window-level representations into a feature sequence. $\text{LN}\left(\cdot\right)$ and $\text{FFN}\left(\cdot\right)$ denote the layer normalization operation and feed-forward network, respectively. $F^{V}\in \mathbb{R}^{T\times D}$ denotes the output features of a video encoder layer, and the output of each former layer will be further fed to the next layer for encoding.

\subsection{Local-Global Iterative Decoder}
Existing methods developed for single-paragraph video grounding ~\cite{depnet, prvg, hscnet} either encode the language query into a single global embedding~\cite{prvg} causing too much information loss, or employ self-attention on the complete multimodal sequence of all text features and video features~\cite{hscnet}, which incurs prohibitive resource cost thus is unsuitable in the multi-paragraph scenario. In this work, we explore a novel way to model the local-global query structures and cross-modal correspondences by iteratively reasoning about the local subparagraph features and global paragraph features.

To begin with, we first utilize a pre-trained RoBERTa~\cite{liu2019roberta} model to obtain the token-level language features from the $i$-th input paragraph, i.e., $F_i^L\in \mathbb{R}^{N_i^L\times D}$, where $N_i^L$ is the total number of language tokens in the $i$-th input paragraph. Afterwards, we jointly utilize a paragraph encoder and a subparagraph extractor to efficiently model the intrinsic local and global structures of the long text inputs, as shown in Figure~\ref{fig:framework}. Concretely, we first embed all the token-level features within a paragraph into a global query feature $Q_i^g\in \mathbb{R}^{D}$ by an average-pooling operation. Then, we exploit $E$ learnable vectors $O^S\in \mathbb{R}^{E\times D}$ to extract the important subparagraph representations under the semantic guidance of $Q_i^g$ as follows:
\vspace{-1mm}
\begin{equation}
    Q_i^{\ell} = \text{Cross-Attention}\left(\text{LN}\left((Q_i^gW_1 + O^SW_2)\right), F_i^L, F_i^L\right)
    \vspace{-1mm}
\end{equation}
where $W_1\in \mathbb{R}^{D\times D}$ and $W_2\in \mathbb{R}^{D\times D}$ are learnable projection matrices and $\text{LN}\left(\cdot\right)$ is the layer normalization operation. $Q_i^{\ell}\in \mathbb{R}^{E\times D}$ is the extracted subparagraph features that can be adaptively learned to represent meaningful local semantics specific to each paragraph for enhancing the cross-modal reasoning abilities within the paragraphs. Note that the number of extracted subparagraph features is typically small and the computation process will be efficient.

After obtaining the local subparagraph features and global paragraph features, we construct an iterative local-global reasoning process where each iteration involves intra-level reasoning, cross-modal reasoning and cross-level reasoning. Firstly, we conduct intra-level reasoning by employing two self-attention layers respectively within each window of local queries $Q_i^{\ell}$ and within all global queries $Q^{g}$. Afterwards, we achieve cross-modal reasoning by extracting relevant information from $F^{V}$ to $Q^{\ell}$ and $Q^{g}$ by cross-attention layers. Then, we conduct cross-level reasoning also by two cross-attention layers, i.e., one is for extracting information from $Q^{g}$ to $Q^{\ell}$ and the other one is for extracting information from a window of local queries $Q_i^{\ell}$ to the corresponding global query $Q_i^{g}$. The cross-interacted features will serve as the output features of each decoder layer and are fed to the next layer for iterative decoding. Finally, the local and global output features, i.e., $Q_i^{\ell}$ and $Q_i^{g}$ of the last decoder layer are concatenated and fed to an MLP predictor to obtain the central timestamp $\hat{t}_c^{i}$ and duration $\Delta \hat{t}^{i}$ of the target interval for the $i$-th paragraph query. Then the temporal boundaries $\mathcal{\hat{T}}_i = (\hat{t}_s^{i}, \hat{t}_e^{i})$ can be calculated as $\hat{t}_s^{i}=\hat{t}_c^{i}-\frac{\Delta \hat{t}^{i}}{2}, \hat{t}_e^{i}=\hat{t}_c^{i} + \frac{\Delta \hat{t}^{i}}{2}$.
\vspace{-2.5mm}
\subsection{Model Training}
We train our model with a localization loss $\mathcal{L}_{loc}$ and an attention loss $\mathcal{L}_{att}$ which are formulated as follows:
\begin{align}
   \mathcal{L}_{loc} &= \frac{1}{N}\sum_{i=1}^{N} \left [ \frac{1}{\lambda_1} \mathcal{L}_{l1}(\mathcal{\hat{T}}_i, \mathcal{T}_i)+ \mathcal{L}_{GIoU}(\mathcal{\hat{T}}_i, \mathcal{T}_i) \right ], \\
   \mathcal{L}_{att} &= -\frac{1}{N}\sum_{i=1}^{N}\text{log}\left (\sum_{j=1}^{T}m_{ij}\cdot a_{ij} \right )
\end{align}
where $\mathcal{L}_{l1}$ and $\mathcal{L}_{GIoU}$ are L1 and GIoU~\cite{giou} losses, respectively. $\mathcal{\hat{T}}_i$ is the predicted time span for the $i$-th query and $\mathcal{T}_i$ is the ground-truth.  $\mathcal{L}_{att}$ is an attention loss on the global query features. The term $a_{ij}$ indicates the attention weights between the $i$-th global query feature and the $j$-th video feature, while $m_{ij}$ is an indicator that takes 1 if the $j$-th video feature is inside the ground-truth interval of the $i$-th query, and 0 otherwise. This loss explicitly encourages the model to learn higher attention weights between text queries and visual elements that are correlated. In total, the training loss is defined as the weighted sum of the above two losses as $\mathcal{L} = \lambda_{1} \mathcal{L}_{loc} + \lambda_{2} \mathcal{L}_{att}$, where $\lambda_{1}$ and $\lambda_{2}$ are the hyper-parameters to balance these two different kinds of losses.

\section{Experiments}
In this section, we illustrate our experimental setup and results for verifying and analyzing the effectiveness of our proposed method.
\vspace{-1mm}
\subsection{Experimental Setup}
\noindent\textbf{Evaluation metrics.} For each query in the synopsis, we calculate the temporal Intersection over Union (IoU) between the predicted time span $\left [\hat{t}_s, \hat{t}_e\right ]$ and the ground-truth time span $\left [t_s,t_e\right ]$. Following previous video grounding methods~\cite{tall_charades, didemo}, we adopt two kinds of metrics to evaluate the performance: 1) mean IoU (mIoU) metric: average temporal IoU score calculated over all queries in the dataset; 2) IoU@$\theta$ metric: the proportion of queries with a temporal IoU score higher than $\theta$, here we use $\theta \in \{0.3, 0.5, 0.7\}$.

\noindent\textbf{Implementation details.} Our proposed method is implemented by PyTorch. The pre-extracted SlowFast, CLIP, and OCR features are aligned at sequence dimension and concatenated at channel dimension as the video feature input. We use a pre-trained RoBERTa~\cite{liu2019roberta} model to extract OCR features at each timestamp. The loss weights are set as $\lambda_{1}=1, \lambda_{2}=0.2$. For data augmentation, we choose to randomly shuffle the order of paragraphs in the same synopsis by a probability of $p$ during training and $p=\text{max}(0, 1-\frac{T_i}{T_{max}})$, where $T_i$ is the index of the current training epoch and $T_{max}$ is set to 20. We adopt a local window length $M$ of 25 for our video encoder. The number of layers for video encoder and query decoder are set as 2 and 3, respectively. Our model is trained on 4 NVIDIA Tesla V100 GPUs by Adam~\cite{adam} optimizer using a learning rate of 0.0001 and batch size of 16 for a total of 50 epochs within one day.

\subsection{Experimental Results}
\begin{table}[t]
    \centering
    \renewcommand{\arraystretch}{1.3}
    \caption{Comparison results with state-of-the-art methods on multi-paragraph video grounding in SynopGround.}
    \vspace{-3mm}
    \resizebox{!}{15mm}{
    \begin{tabular}{c|c|c|c|c}
        \hline
        Method & mIoU & IoU@0.3 & IoU@0.5 & IoU@0.7\\
        \hline
        Human & 85.1 & 97.3 & 92.5 & 85.0 \\
        Random & 7.3 & 8.3 & 3.2 & 0.8 \\
        \hline
        DepNet~\cite{depnet} & 30.7 & 47.2 & 28.7 & 12.8 \\
        PRVG~\cite{prvg} & 34.7 & 52.7 & 29.3 & 10.5 \\
        LGMR (Ours) & \textbf{44.4} & \textbf{67.9} & \textbf{46.7} & \textbf{21.8} \\
        \hline
    \end{tabular}}
    \label{tab:main_res}
    \vspace{-4mm}
\end{table}
\begin{table}[t]
    \centering
    \renewcommand{\arraystretch}{1.3}
    \caption{Evaluation on the effect of different features.}
    \vspace{-3mm}
    \resizebox{!}{19.5mm}{
    \begin{tabular}{ccc|cccc}
        \hline
        SlowFast & CLIP & OCR & mIoU & IoU@0.3 & IoU@0.5 & IoU@0.7 \\
        \midrule
        \checkmark & $\times$ & $\times$ & 39.1 & 60.5 & 39.3 & 17.1 \\
        $\times$ & \checkmark & $\times$ & 39.8 & 61.7 & 40.0 & 16.5 \\
        $\times$ & $\times$ & \checkmark & 41.7 & 64.3 & 43.4 & 17.7 \\
        \checkmark & \checkmark & $\times$ & 41.0 & 62.8 & 41.7 & 18.1 \\
        \checkmark & $\times$ & \checkmark & 43.0 & 67.4 & 43.8 & 19.4 \\
        $\times$ & \checkmark & \checkmark & 43.8 & 67.1 & 46.0 & 21.3 \\  
        \checkmark & \checkmark & \checkmark & \textbf{44.4} & \textbf{67.9} & \textbf{46.7} & \textbf{21.8} \\
        \hline
    \end{tabular}}
    \label{tab:ablation1}
    \vspace{-5mm}
\end{table}
\noindent\textbf{Performance Comparison.} As shown in Table~\ref{tab:main_res}, we evaluate the performance of our proposed LGMR on the challenging multi-paragraph video grounding task and compare it with the existing state-of-the-art methods DepNet~\cite{depnet} and PRVG~\cite{prvg}. DepNet is the baseline method proposed for multi-sentence video grounding, and PRVG is a concise and effective method based on DETR-like architectures~\cite{detr}. For a fair comparison, all reported methods employ the same features as ours. The comparison results in Table \ref{tab:main_res} demonstrate that our proposed model achieves the best performance and outperforms others by a significant margin, which validates the effectiveness of the proposed LGMR method for addressing MPVG.

\noindent\textbf{Impact of different features.} To investigate the effect of different features, we conduct experiments with various combinations of SlowFast, CLIP, and OCR features. As shown in Table~\ref{tab:ablation1}, we observe that using a single kind of features already yields satisfactory performance. Specifically, using the SlowFast, CLIP, or OCR features alone is able to produce an mIoU of $39.1\%$, $39.8\%$, and $41.7\%$, respectively. We notice that the CLIP features and OCR features are more helpful than the SlowFast features, which might be because 1) CLIP is pre-trained on large-scale image-text pairs, which makes it generalize better to the downstream task of video-language grounding. 2) The OCR features encode rich character-related and dialogue information, which is important for understanding the story plots in the narrative video. Additionally, we can see that the model using all three features together achieves the best performance on all metrics, showing that different kinds of features convey complementary information of the video content for language grounding.

\noindent\textbf{Effect of the local-global query modeling.} As shown in Table~\ref{tab:ablation_query_decoder}, we conduct detailed experiments to verify our proposed idea to model and reason the local-global structures of long queries. First, the model using only local queries for the cross-modal decoding process achieves a significantly lower performance compared to our final model. The reason is that only considering intra-query semantics neglects the rich contextual relationships among multiple correlated queries, while understanding the contexts is crucial for the multi-paragraph video grounding problem. Secondly, we observe significant gains in performance when jointly modeling the local and global structures of the long text inputs during decoding, showing the importance of our local-global query modeling.

\noindent\textbf{Ablation studies on design choices.} To further validate the rationality of our proposed model, we conduct ablation experiments on the designs of local-global temporal attention and cross-modal attention loss, as shown in Table~\ref{tab:ablation_component}. Firstly, we compare our model performance with that of a variant model where the local-global encoder is replaced by a vanilla full attention encoder~\cite{attention}. The result suggests that our local-global encoder performs better in both accuracy and efficiency for long video inputs. Besides, we remove $\mathcal{L}_{att}$ and observe severe degradation in model performance. This highlights the importance of explicitly guiding the model to associate and align correlated visual and textual features.
\begin{table}[t]
    \centering
    \renewcommand{\arraystretch}{1.3}
    \caption{Effect of the local-level modeling, global-level modeling, and cross-level reasoning in the iterative query decoder.}
    \vspace{-3mm}
    \resizebox{!}{11mm}{
    \begin{tabular}{ccc|cccc}
        \hline
        Local & Global & Cross & mIoU & IoU@0.3 & IoU@0.5 & IoU@0.7 \\
        \midrule
        \checkmark & $\times$ & $\times$ & 34.5 & 53.1 & 32.4 & 13.5 \\
        \checkmark & \checkmark & $\times$ & 42.8 & 66.6 & 44.2 & 19.0 \\
        \checkmark & \checkmark & \checkmark & \textbf{44.4} & \textbf{67.9} & \textbf{46.7} & \textbf{21.8} \\
        \hline
     \end{tabular}}
    \label{tab:ablation_query_decoder}
    \vspace{-3.5mm}
\end{table}

\begin{table}[t]
    \centering
    \renewcommand{\arraystretch}{1.25}
    \caption{Ablation studies on the proposed model designs. The GFLOPs measures computation complexity of the encoder.}
    \vspace{-3mm}
    \resizebox{!}{15mm}{
    \begin{tabular}{c|c|c|c|c}
        \hline
        Encoder & mIoU & GFLOPs & IoU@0.5 & IoU@0.7 \\
        \hline
        Vanilla Full & 42.5 & 12.6 & 45.1 & 20.5  \\
        Local-Global & \textbf{44.4} & \textbf{9.6} & \textbf{46.7} & \textbf{21.8} \\ 
        \hline
        Loss & mIoU & IoU@0.3 & IoU@0.5 & IoU@0.7 \\
        \hline
        $\mathcal{L}_{loc}$ & 26.7 & 40.2 & 16.6 & 4.8 \\
        $\mathcal{L}_{loc}$ and $\mathcal{L}_{att}$ & \textbf{44.4} & \textbf{67.9} & \textbf{46.7} & \textbf{21.8} \\
        \hline
    \end{tabular}}
    \label{tab:ablation_component}
    \vspace{-4mm}
\end{table}

\section{Conclusion}
In this work, we present a large-scale dataset for video-language grounding called SynopGround, which consists of over 2800 hours of long narrative videos with human-written synopses and manually annotated timestamps. It is the first video grounding dataset considering both long-form videos and long-text queries, and contains query descriptions conveying both low-level events as well as high-level plots for learning more complex and abstract concepts. We further introduce a challenging Multi-Paragraph Video Grounding (MPVG) task which incorporates long paragraph queries into multi-query video grounding. In addition, we propose a novel Local-Global Multimodal Reasoner (LGMR) to explicitly model the local-global structures of long-term inputs and conduct iterative reasoning within and across the two levels of structures, which can serve as a good starting point to inspire future research.

\begin{acks}
This work was supported partially by the NSFC (U21A20471, U22A2095, 62076260, 61772570), Guangdong Natural Science Funds Project (2023B1515040025, 2022B1111010002), Guangdong NSF for Distinguished Young Scholar (2022B1515020009), Guangdong Provincial Key Laboratory of Information Security Technology (2023B1212060026), Guangzhou Science and Technology Plan Project (202201011134) and GACC-Research Projects (2022HK135).
\end{acks}
\bibliographystyle{ACM-Reference-Format}
\balance
\bibliography{ref}

\clearpage
\appendix

\section{Supplementary Material}
In this supplementary material, we provide more information including additional experimental results, more implementation details and qualitative analysis based on visualization.

\subsection{Additional Experimental Results}
In this part, we present some additional comparison and ablation results, to further investigate and validate the effectiveness of our proposed designs for the baseline model LGMR.

\noindent\textbf{Comparison on MAD dataset.} In Table ~\ref{tab: mad_comparison}, we show the comparison results with other state-of-the-art methods on the 3-min version of MAD dataset, following the setting in~\cite{soldan2022mad}, to verify the superiority of our proposed LGMR method over existing baselines. As can be seen, our method consistently outperforms state-of-the-art multi-query methods in all metrics by a large margin.
\begin{table}[h]
    \vspace{-3.5mm}
    \footnotesize
    \centering
    \caption{Performance comparison on the MAD dataset.}
    \vspace{-3.5mm}
    \begin{tabular}{c|ccccc}
        \hline
        Method & Query Input & R@ 0.1 & R@0.3 & R@0.5 & mIoU \\
        \hline
        DepNet~\cite{depnet} & Multiple & 21.5 & 15.0 & 8.3 & 9.6 \\
        PRVG~\cite{prvg} & Multiple & 37.9 & 15.0 & 5.7 & 12.3 \\
        LGMR (Ours) & Multiple & \textbf{51.7} & \textbf{31.4} & \textbf{14.6} & \textbf{20.9} \\
        \hline
    \end{tabular}
    \label{tab: mad_comparison}
    \vspace{-3.5mm}
\end{table}

\noindent\textbf{Ablation Study on Loss Weights.} As shown in Table~\ref{tab: hyper-parameter}, we actually determined the two loss weights by a grid search. It can be observed that setting $\mathcal{L}_{1}$ to be relatively larger than $\mathcal{L}_{2}$ gives a decently good model performance, and the best choice is $\mathcal{L}_{1}=1.0$ and $\mathcal{L}_{2}=0.2$.
\begin{table}[h]
    \vspace{-3.5mm}
    \footnotesize
    \centering
    \caption{Hyper-parameter search in terms of $\mathcal{L}_{1}$ and $\mathcal{L}_{2}$}
    \vspace{-3.5mm}
    \begin{tabular}{cccccc}
        \hline
        $\mathcal{L}_{1}$ & $\mathcal{L}_{2}$ & R@ 0.3 & R@0.5 & R@0.7 & mIoU \\
        \hline
        1.0 & 1.0 & 64.4 & 42.4 & 17.0 & 41.0 \\
        1.0 & 0.5 & 64.4 & 44.4 & 19.0 & 41.9 \\
        1.0 & 0.2 & \textbf{67.9} & \textbf{46.7} & \textbf{21.8} & \textbf{44.4} \\
        0.5 & 0.2 & 67.5 & 46.7 & 20.5 & 43.7 \\
        \hline
    \end{tabular}
    \label{tab: hyper-parameter}
    \vspace{-3.5mm}
\end{table}

\noindent\textbf{Comparison with Single-Sentence Methods.} For a more comprehensive comparison, we show the comparative results of our LGMR with three single-query state-of-the-arts, i.e., CONE~\cite{cone}, 2D-TAN~\cite{2dtan} and VSLNet~\cite{vslnet}. As shown in Table~\ref{tab: single_comparison}, our LGMR surpasses all single-query methods by a large margin. Note that CONE performs the worse since it is a method directly built on top of pre-trained vision-text models while the vision and text features in our dataset are not pre-aligned. VSLNet is the best-behaved single-query method since it generally models long-term video inputs by a well-designed split-and-concat mechanism.
\begin{table}[h]
    \vspace{-3.5mm}
    \footnotesize
    \centering
    \caption{Comparison with extra single-query baselines.}
    \vspace{-3.5mm}
    \begin{tabular}{c|ccccc}
        \hline
        Method & Query Input & R@ 0.3 & R@0.5 & R@0.7 & mIoU \\
        \hline
        CONE~\cite{cone} & Single & 4.7 & 1.8 & 0.5 & - \\
        2D-TAN~\cite{2dtan} & Single & - & 8.8 & 3.2 & 11.5 \\
        VSLNet~\cite{vslnet} & Single & 45.0 & 30.8 & 18.6 & 32.8 \\
        \hline
        DepNet~\cite{depnet} & Multiple & 47.2 & 28.7 & 12.8 & 30.7 \\
        PRVG~\cite{prvg} & Multiple & 52.7 & 29.3 & 10.5 & 34.7 \\
        LGMR (Ours) & Multiple & \textbf{67.9} & \textbf{46.7} & \textbf{21.8} & \textbf{44.4} \\
        \hline
    \end{tabular}
    \label{tab: single_comparison}
    \vspace{-3.5mm}
\end{table}

\subsection{More Implementation Details}
We provide further implementation details for our proposed baseline model, including feature dimensions, positional encodings, subparagraph extractor, and loss calculation.

\noindent\textbf{Feature Dimensions.} We use pre-extracted SlowFast~\cite{feichtenhofer2019slowfast}, CLIP~\cite{clip}, and OCR features with dimensions of 2304, 768, and 768, respectively. The CLIP and OCR features are first adaptively pooled at the sequence dimension to have the same length as SlowFast features. Then the three types of features are concatenated together at the hidden dimension as the input video features with a hidden dimension of 3840. A fully-connected layer is used to project the input video features to a 512-dimensional video representation for the temporal encoding and query decoding. Likewise, the pre-extracted text features have a hidden dimension of 768, and they are projected by a fully-connected layer to have the same feature dimension of 512. For all transformer layers in the encoders and decoders, the feature dimension is 512. All feed-forward layers have a hidden dimension of 2048 and the number of attention heads is set to 8.

\noindent\textbf{Positional Encodings.} As proposed in the vanilla transformer architecture~\cite{attention}, we adopt a fixed set of high-dimensional sinusoidal embeddings to indicate positional information. The positional embeddings are employed on all transformer layers, including transformer layers used in the local-global temporal encoder and the local-global iterative decoder. Following the designs in DETR~\cite{detr}, we only add positional embeddings with the feature inputs of query projection layers and key projection layers in all attention blocks.

\noindent\textbf{Subparagraph Extractor.} To construct the local subparagraph features from token-level text features for local-global iterative reasoning in our decoder, we adopt a set of learnable vectors $O^{\mathcal{S}}\in \mathbb{R}^{E\times D}$ to represent potential meaningful local semantics in a paragraph and then use one transformer decoder layer to extract useful subparagraph features in an end-to-end manner. Here we empirically set the number of learnable vectors $E$ to be $10$ in all our experiments and the positional embeddings are also only added to the inputs of query projection layers and key projection layers.

\noindent\textbf{Loss Calculation.} We calculate the localization loss $\mathcal{L}_{loc}$ and attention loss $\mathcal{L}_{att}$ for all transformer decoder layers in our local-global iterative decoder, following the common practice in DETR~\cite{detr} series works. Specifically, we feed the output global paragraph features of each decoder layer to an MLP predictor to predict the starting and ending timestamps. Note that we use a shared layer normalization module to normalize the output features of all layers before using them to predict the temporal interval corresponding to each paragraph query. During training, $\mathcal{L}_{loc}$ is calculated on predictions produced by each decoder layer. Similarly, our attention loss is calculated on the temporal attention weights produced by each decoder layer during training. For testing, we only take the timestamp predictions from the last decoder layer of the model.

\subsection{Qualitative Analysis}
In this section, we aim to conduct qualitative analysis based on the visualization results, which can give a more intuitive understanding of our multi-paragraph video grounding dataset and model.
\subsubsection{Visualization Results}
\label{sec:qualitative}
\begin{figure*}[p]
   \centering
   \includegraphics[width=1.0\linewidth, height=21cm]{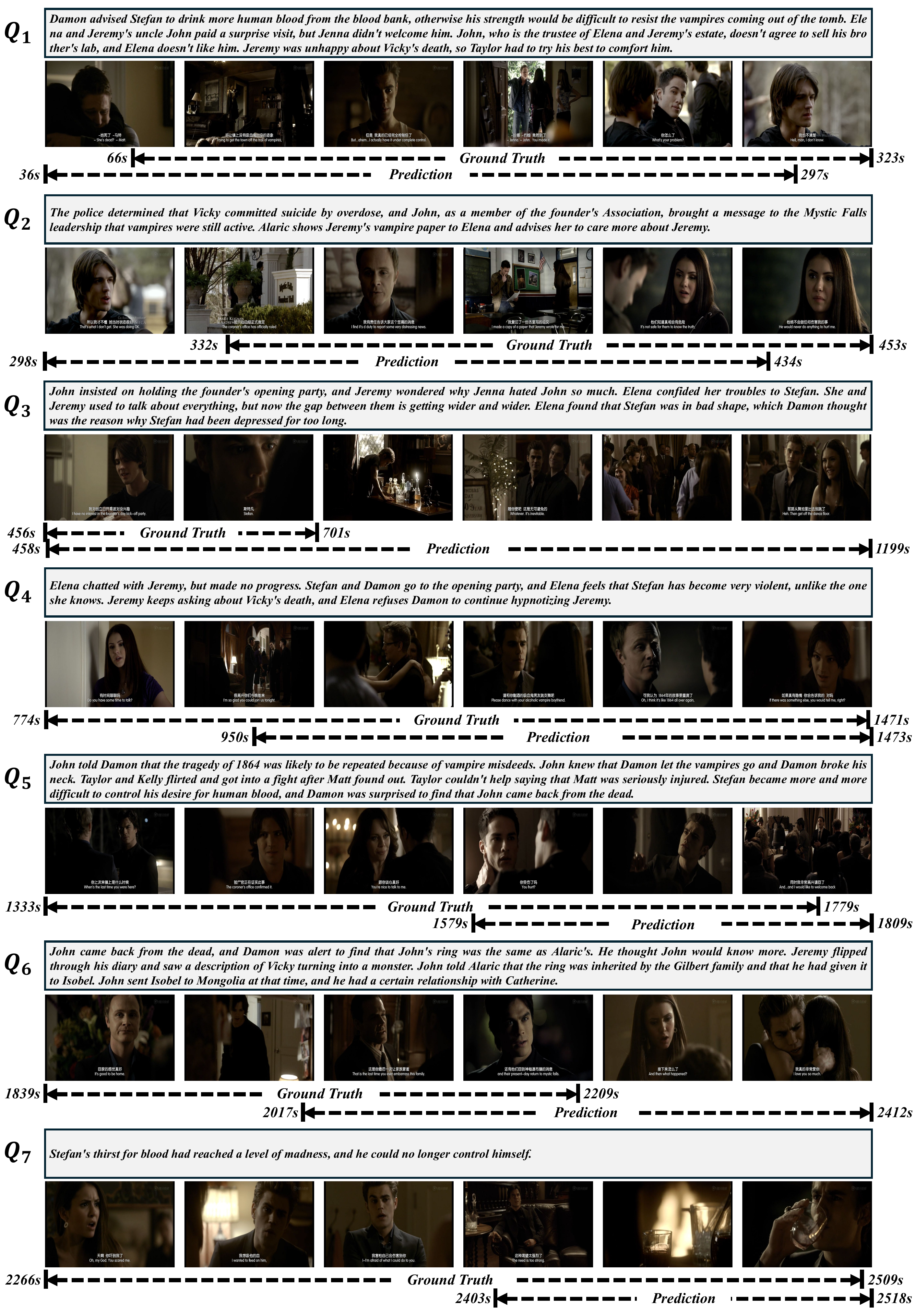}
   \vspace{-9mm}
   \caption{Visualization results of the model predictions and ground truth for multi-paragraph video grounding in SynopGround dataset. This example is selected from the test set and the video is the 18-th episode of the TV Drama \textit{Vampire Diaries Season 1}. Due to space limitation, we uniformly sample frames from the temporal interval that encloses both the model predictions and ground truth for better visual presentation. (Best viewed on screen when zoomed in)}
   \label{fig:visualized_results}
\end{figure*}
First of all, we visualize and present a complete video-synopsis pair from the test set, as shown in Figure~\ref{fig:visualized_results}. This synopsis is composed of seven paragraph queries and most of them are very lengthy and complex. In addition, it can be seen that these paragraph queries typically contain multiple sentences that describe a variety of concepts at different levels of abstraction. For example, there are some abstract and concise expressions like ``confided her troubles to'' that summarize a long character conversation and convey the abstract concept ``troubles'' that may need a certain level of contextual reasoning capability to acquire an accurate understanding of it. Also, there are some concrete and detailed descriptions like ``Jeremy flipped through his diary and saw a description of Vicky turning into a monster'', which requires to comprehend rich visual details presented in the video content for multimodal understanding. As a result of the above characteristics, jointly conducting contextual understanding of the video storylines and comprehensive perception of the visual details in each paragraph poses a crucial challenge for the video-language grounding models to overcome. Note that the target moments are also lengthy with a duration of several minutes, which requires models to effectively capture the more complex temporal structures of the video moments while retaining the ability of memorizing long-term visual contexts for better reasoning across multiple moments.

In Figure~\ref{fig:visualized_results}, we also compare our model's predicted temporal intervals with the ground-truth timestamps to intuitively demonstrate the abilities of our multi-paragraph video grounding model. Overall, our model can make predictions close to the ground truth and correctly determine most of the temporal boundaries for the target video moments described by the given paragraph queries, although in some cases the boundary locations predicted by the model may not be very precise. To conduct a more detailed analysis of the model predictions, we further present the text content of each paragraph query in the synopsis and visualize some frames from the video moments corresponding to these queries, as illustrated in Figure~\ref{fig:visualized_results}. On the one hand, we can see the model is able to successfully predict the temporal intervals that have a high degree of overlap with the ground truth for the first two and the fourth paragraph queries, i.e., $Q_1$, $Q_2$ and $Q_4$. In these cases, the paragraph queries are complicated and lengthy while containing rich complex concepts such as ``Jeremy keeps asking about Vicky's death, and Elena refuses Damon to continue hypnotizing Jeremy" and ``Elena and Jeremy's uncle paid a surprise visit, but Jenna didn't welcome him". Understanding these complex concepts requires the model to have a strong ability to associate a broad range of textual semantics in the paragraphs with the dialogue information as well as the visual activities in the video for precise temporal grounding.

On the other hand, we also observe some cases where the predictions are not very accurate. For instance, one of the two predicted temporal boundaries is accurate while the other one deviates from the ground truth by a considerable margin for the third, fifth and seventh paragraphs, i.e., $Q_3$, $Q_5$ and $Q_7$. We analyze them to find potential reasons case by case. For the third paragraph query in the synopsis, the model predicts a very accurate starting timestamp but predicts a much later ending timestamp in the video. In this case, localizing the ending timestamp is closely relevant to finding out the dialogue information between the drama characters referred to by the description ``Elena found that Stefan was in bad shape, which Damon thought was the reason why Stefan had been depressed for too long''. Furthermore, we find the model incorrectly predicts the ending timestamp to be around the $20$-th minute in the video, where a salient visual activity concerning the physical conflicts between two characters is located. This might indicate our model's deficiency in resisting the distraction from irrelevant salient visual activities. For the fifth paragraph query, its predicted starting timestamp is later than the ground-truth starting timestamp. In this example, we find that the incorrectly predicted starting time is actually corresponding to the third sentence in this paragraph, i.e., ``Taylor and Kelly flirted and got into a fight after Matt found out'', which means the model has missed the information associated with the first two sentences in the paragraph during video-language grounding. This phenomenon highlights the importance of fully understanding all the necessary detailed information contained in the long-term textual content of the paragraph queries and our model still needs to be improved in this aspect.

For the seventh paragraph query, we notice that the model predicts the starting timestamp to be around the point where the two characters' dialogue mentions ``the need is too strong'' which is directly related to the key word ``thirst'' in the given paragraph. The model can only make its prediction for this case by considering the above kind of simple correlations between the video content and query semantics, thus causing an inaccurate starting boundary. Actually, the described character shows a very struggling and painful expression and body movements at the ground-truth starting time, which implicitly indicates the start of the video content specified by the query. However, the model fails to perceive such subtle human facial expressions and body movements to associate them with the plot contexts for predicting the starting boundary, which suggests this kind of ability needs to be further developed in future research. Last but not least, we also find the overall position of the predicted temporal interval of the sixth paragraph query $Q_6$ is shifted a bit to the right at the time axis. In this case, we find that the starting time given by the model is very close to the moment corresponding to the third sentence in the paragraph, i.e., ``Jeremy flipped through his diary and saw a description of Vicky turning into a monster'', which implies the model may miss the query information of the first two sentences in the paragraph. For the ending time, we find that localizing the ground-truth boundary requires to capture the short-term dialogue information implicitly corresponding to the query description ``John told Alaric that the ring was inherited by the Gilbert family and that he had given it to Isobel'', while localizing a piece of short-term information from a long-term input is challenging and the model can be struggling to handle such cases.

In conclusion, the temporal intervals predicted by our baseline model can decently overlap with the ground truth in various cases, demonstrating the model's ability to associate most concepts across visual and linguistic modalities. However, our model still struggles to predict very precise temporal boundaries in some challenging cases that demand deep understanding and complex reasoning of the story's global context and crucial nuances. This points to an important direction for future work to develop stronger models that can better integrate global context and local details across modalities and conduct complex reasoning in a long contextual scope for better multi-paragraph video grounding.

\subsubsection{Analysis on Attention Weights}
\begin{figure*}[p]
   \centering
   \includegraphics[width=1.0\linewidth, height=21.5cm]{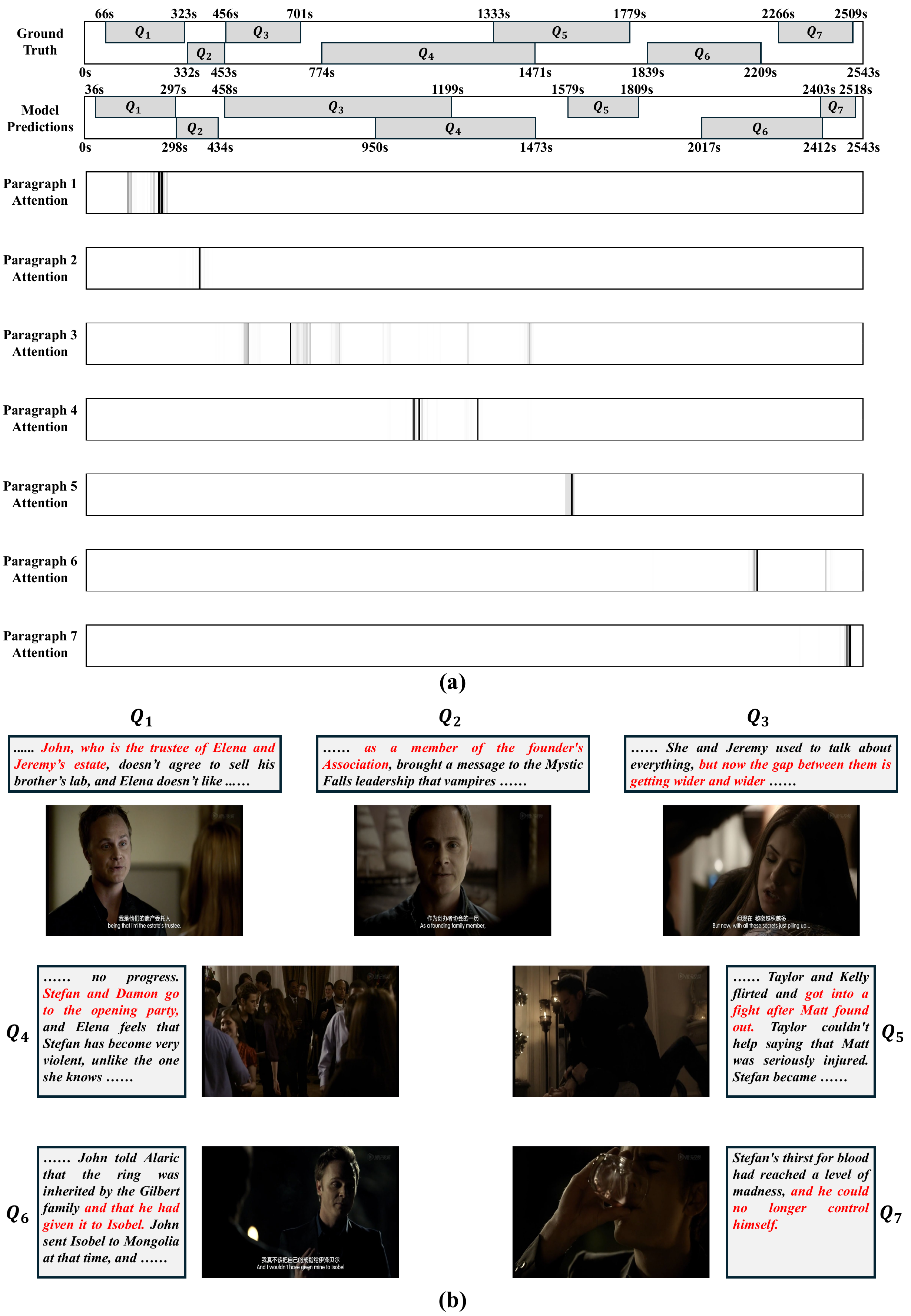}
   \vspace{-9mm}
    \caption{(a) Visualization on the paragraph-to-video attention weights from the last decoder layer. (b) Visualization on frames around the attention peak of the paragraph queries. Red text denotes the part of content within each paragraph query in the synopsis that is directly related to the visual frame located around an attention peak. (Best viewed on screen when zoomed in)}
   \label{fig:visualized_attention}
\end{figure*}
\begin{figure*}[h] 
   \centering
   \includegraphics[width=1.0\linewidth]{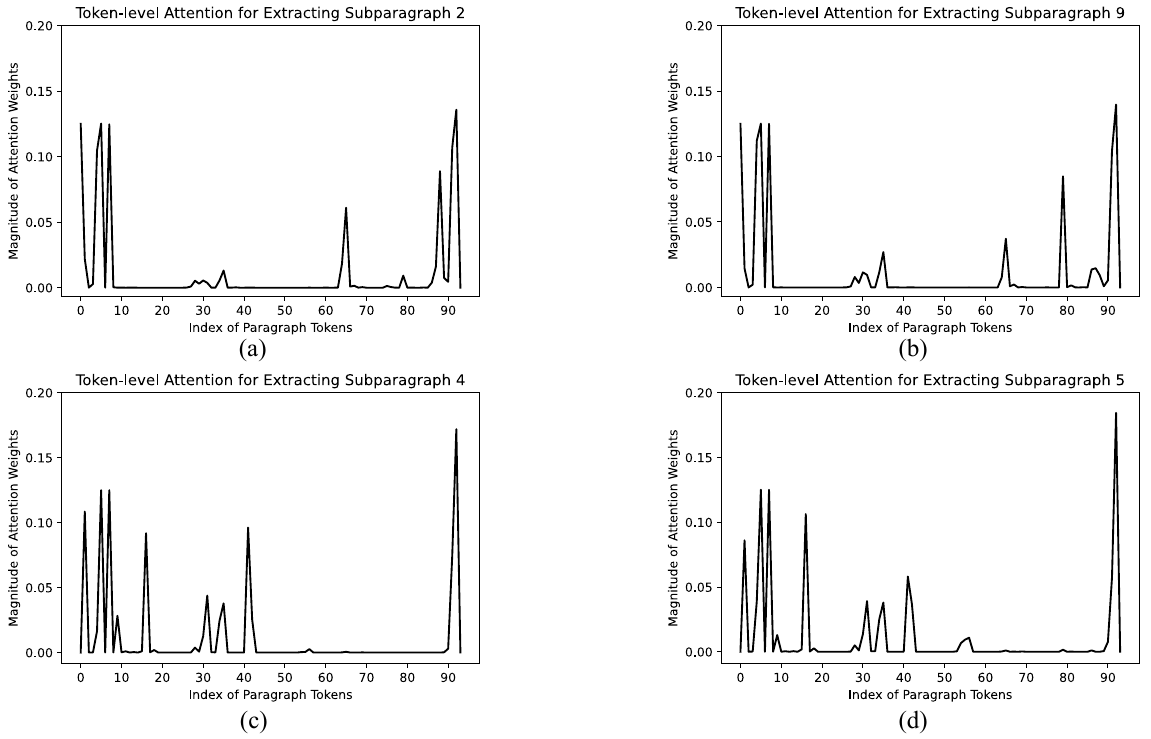}
   \vspace{-5.3mm}
   \caption{Visualization results of the token-level attention weights for different local subparagraph representations. (a), (b), (c) and (d) respectively illustrate the attention weights from subparagraph 2, 9, 4 and 5 in the subparagraph extractor.}
   \label{fig:visualized_subparagraph}
   \vspace{-2.5mm}
\end{figure*}

\noindent\textbf{Paragraph-to-Video Attention Weights.} In Figure~\ref{fig:visualized_attention}, we visualize the learned temporal attention weights from the last layer of the query decoder for the sample discussed in Section~\ref{sec:qualitative}. To make a clearer visualization presentation, we additionally show the predicted timestamps of all paragraph queries in this sample along with the corresponding ground-truth labels in the upper part of Figure~\ref{fig:visualized_attention} (a). As we can see, both of the model's final predictions and attention weights obviously follow a consistent temporal order with the ground-truth temporal intervals and the model's predictions are highly correlated with the temporal positions where higher attention weights occur. This phenomenon intuitively suggests that the learned correlation between language queries and video content is crucial for achieving accurate temporal event localization. Encouraging high attention weights for relevant query and video elements is therefore beneficial for video-language grounding, which has also been quantitatively verified by the remarkable effect of the cross-modal attention loss according to our manuscript. In particular, we also observe that for the third paragraph query $Q_3$, there are some temporal positions far away from the target moment that are spuriously attended by the model, which directly leads to a considerably delayed ending timestamp predicted by the model. Upon manually reviewing the corresponding video content that is spuriously attended by the model, we find that the main reason for the inaccurate prediction in this case lies in the model's inability to correctly understand ``She and Jeremy used to talk about everything''. In fact, the mistakenly attended frames are about the two characters talking with each other along the riverside, while the model incorrectly associates such content with ``used to talk'', leading to inaccurate boundaries.

To more comprehensively understand the attention patterns of the decoder, we select video frames that are located around the temporal attention peaks of different paragraph queries and visualize them in Figure~\ref{fig:visualized_attention} (b). Overall, we observe that these frames with high attention weights from the paragraph queries are consistently correlated with certain descriptions in the corresponding paragraph, as shown by the red text in Figure~\ref{fig:visualized_attention} (b). Specifically, the dialogue information of the video content can be viewed as directly correlated with some part of the query content for $Q_1$, $Q_2$, $Q_3$ and $Q_6$. In these cases, characters in the frames are talking about crucial information mentioned by the query. For example, the character is introducing his identity as a ``founding family member'' in the visualized frame of $Q_2$, while this information is exactly mentioned in the second query by ``as a member of the founder's Association''. In addition to that, there are also cases where the dialogue information in the visualized frame is implicitly related to the query text. For instance, for the third query $Q_3$, the character is saying ``with all these secrets just piling up''. This dialogue does not explicitly mention information about the ``gap'' but actually implies the gap between the two characters is becoming wider, which is described in the query as ``but now the gap between them is getting wider and wider''. Furthermore, there are also cases where the visualized frames present the visual activities referred to by the corresponding query content, such as the situations in $Q_4$, $Q_5$ and $Q_7$. Concretely, characters are seen fighting in the frame relevant to $Q_5$, which is exactly described by the query as ``got into a fight after Matt found out''. Particularly, the character is shown to be struggling inside and finally ends up drinking a cup of blood on the table, and this visual activity actually corresponds to the description ``and he could no longer control himself'' in the seventh query. In summary, we find that our model has the ability to find cross-modal correlations between query descriptions and the video content, regardless of whether information from different modalities is correlated explicitly or implicitly through character dialogues or visual activities.

\noindent\textbf{Subparagraph-to-Token Attention Weights.} In Figure~\ref{fig:visualized_subparagraph}, we further visualize the subparagraph-to-token attention weights in the query decoder to better analyze the local-level structure modeling in our local-global reasoning process. As presented, the different subparagraph features successfully learn to attend over different parts of the language tokens in the paragraph query. Intuitively, the attention weights from different subparagraphs can be roughly grouped into two patterns, i.e., the pattern shared by (a) and (b) and the pattern shared by (c) and (d), while different subparagraphs belonging to the same pattern still show a certain level of diversity. Moreover, the local semantics captured by different patterns of subparagraph representations are highly complementary to each other. For example, the attention weights in (c) mainly focus on language tokens at the front of the paragraph while the attention weights in (a) focus more on language tokens at the end of the paragraph. The complementary information contained by multiple subparagraphs helps the model to efficiently extract local semantic details. Therefore, adaptively extracting subparagraph features is an effective way to construct the local-global cross-modal reasoning process regarding the long-term multimodal inputs.

\end{document}